\documentclass[10pt,twocolumn,letterpaper]{article}

\usepackage{cvpr}              

\usepackage{graphicx}
\usepackage{amsmath}
\usepackage{amssymb}
\usepackage{booktabs}
\usepackage{subcaption}
\usepackage{array,multirow}
\usepackage{booktabs}
\usepackage{graphicx}
\usepackage[table]{xcolor}
\usepackage{pifont}
\newcommand{\cmark}{\ding{51}}%
\newcommand{\xmark}{\ding{55}}%

\usepackage{dblfloatfix}

\usepackage[pagebackref,breaklinks,colorlinks]{hyperref}

\usepackage[capitalize]{cleveref}
\crefname{section}{Sec.}{Secs.}
\Crefname{section}{Section}{Sections}
\Crefname{table}{Table}{Tables}
\crefname{table}{Tab.}{Tabs.}

\def\cvprPaperID{*****} 
\def\confName{CVPR}
\def\confYear{2022}

\begin{document}

\title{Unsupervised Class Generation to Expand Semantic Segmentation Datasets}

\author{Javier Montalvo\\
\and
Álvaro García-Martín\\
\and
Pablo Carballeira\\
\and
Juan C. SanMiguel\\}
\maketitle

\begin{abstract}
Semantic segmentation is a computer vision task where classification is performed at a pixel level. Due to this, the process of labeling images for semantic segmentation is time-consuming and expensive. To mitigate this cost there has been a surge in the use of synthetically generated data --usually created using simulators or videogames-- which, in combination with domain adaptation methods, can effectively learn how to segment real data. Still, these datasets have a particular limitation: due to their closed-set nature, it is not possible to include novel classes without modifying the tool used to generate them, which is often not public. Concurrently, generative models have made remarkable progress, particularly with the introduction of diffusion models, enabling the creation of high-quality images from text prompts without additional supervision.

In this work, we propose an unsupervised pipeline that leverages Stable Diffusion and Segment Anything Module to generate class examples with an associated segmentation mask, and a method to integrate generated cutouts for novel classes in semantic segmentation datasets, all with minimal user input. Our approach aims to improve the performance of unsupervised domain adaptation methods by introducing novel samples into the training data without modifications to the underlying algorithms. With our methods, we show how models can not only effectively learn how to segment novel classes, with an average performance of 51\% IoU, but also reduce errors for other, already existing classes, reaching a higher performance level overall.
\end{abstract}

\section{Introduction}

Semantic segmentation is a fundamental task in computer vision, as it enables spatial location and classification of elements in a given image.  This process is crucial for various applications, including autonomous driving, medical imaging, and robotics. However, creating high-quality, pixel-level annotated datasets for training effective semantic segmentation models is labor-intensive and costly \cite{Cordts2016Cityscapes}.

Researchers have increasingly turned to synthetic data as a viable alternative for training semantic segmentation models to address these challenges. Datasets like Synthia\cite{Ros2016synthia} or GTAV\cite{Richter_2016_ECCV} use a simulation tool and a videogame respectively to generate synthetic images with semantic segmentation labels and provide many advantages compared to real datasets,  including perfect ground truth annotations, controlled environments, and the ability to generate large-scale datasets efficiently, without unaffordable costs. However, synthetic data also has its drawbacks: the inherent domain gap between synthetic and real-world images often leads to suboptimal performance when models trained on synthetic data are applied to real-world scenarios\cite{10128983}. Moreover, existing synthetic datasets often suffer from limitations such as a reduced set of object classes or limited variability within certain categories compared to real-world data. These constraints can hinder the generalization capabilities of models trained on synthetic data.

Recently, the integration of natural-language-processing methods in computer vision tasks has enabled remarkable advancements in the field, particularly in the field of image synthesis with the introduction of methods like Stable Diffusion\cite{rombach_high-resolution_2022}. However, these synthesized images have some limitations when used for semantic segmentation, mainly due to noisy labels. 

In light of these developments, we propose a novel method that leverages Stable Diffusion\cite{rombach_high-resolution_2022} along Segment-Anything-Model \cite{kirillov2023segment} (which is an algorithm capable of obtaining accurate semantic masks using spatial priors of the target object), to successfully exploit the generative capabilities of Stable Diffusion for creating objects and their semantic masks. Moreover, we prove its applicability by expanding currently existing synthetic datasets with additional classes generated with this method, without architectural modifications of the semantic segmentation methods. We show that these novel classes can be effectively learnt in unsupervised domain adaptation pipelines, reaching a level of performance similar to other existing classes in the dataset.

This paper is organized as follows: First, we introduce some relevant works, then we introduce our generation pipeline and combination method, followed by an experiments section where we prove the effectiveness of our method.

\section{Related Work}
\subsection{Generating Synthetic Data}
Synthetic datasets are a valuable tool for semantic segmentation pipelines, with advantages such as reduced gathering costs and efficient data generation, but also drawbacks such as reduced performance on our target, real data, compared to the performance on synthetic data\cite{10128983}.

\paragraph{Synthetic environments}
Synthetic environments have been used for data gathering and testing in different tasks, as they provide a simple and effective playground that not only proves to be cost-efficient but also enables the generation of data that may be dangerous or too costly to collect.

In recent years, synthetic data has usually been generated with tools designed specifically for the task of data simulation and gathering. CARLA \cite{Dosovitskiy17CARLA} or LGSVL \cite{rong2020lgsvl}, are widely used tools designed for simulating autonomous driving tasks, although there are tools for different purposes, such as AirSim \cite{shah2018airsim} for simulating drone navigation. There are also commonly used synthetic datasets like Synthia\cite{Ros2016synthia} or SynLiDAR\cite{xiao2022transfer}, generated with unpublished tools. Additionally, modified videogames can also become a useful source for synthetic data: The GTA Dataset \cite{Richter_2016_ECCV}, was captured using the Grand Theft Auto V video game, and although it required manual label curation, it has become a staple dataset in the field, as it leverages the variability and image quality of a multi-million dollar entertainment project to obtain useful ground-truth labels for different tasks.

Yet, creating these tools and generating these datasets have two notable limitations:  extensive human labor with specific knowledge is required to build these tools, and a limited class variability toppled by the amount of 3D objects embedded into these tools. 

\paragraph{Diffusion models}
Diffusion models have disrupted the field of generative artificial intelligence, as they can generate realistic imaging just from a set of text prompts, without specific tuning and all of this while ensuring high intra and inter-class variability. Yet, they have some limitations that restrict their usage when generating synthetic data for semantic segmentation. The extraction of pixel-accurate semantic labels from images generated with diffusion models is still an open problem, as it has to solve two particular challenges: correctly identifying the class of the depicted object, and the pixel-perfect localization of the object in the image, as the noise introduced by label-misalignment can result in performance degradation on the semantic segmentation models. Some works try to exploit the capabilities of Stable Diffusion generating synthetic images for semantic segmentation: DatasetDM \cite{wu2023datasetdm} introduces a generation model with capabilities to generate images and their ground-truth annotations relying on a segmentation decoder to achieve image-mask alignment. More recently \cite{jia2023dginstyle} relies on ControlNet to generate a diverse dataset of street scenes with consistent image-label alignment, but both methods show limitations with pixel-misalignments and label preservation.

\subsection{Exploiting Synthetic Data}
Although using synthetic environments to generate data has many advantages, it also has some drawbacks and limitations\cite{10128983}. They often have a reduced variability compared to real-world images, due to a limited amount of 3D objects, for example, a CARLA\cite{Dosovitskiy17CARLA} has 18 different car models, but there are thousands in the real world. Also, these environments often lack realism, and seeking realism can increase the development costs of the tool quickly to the point where it may surpass the cost of gathering and labeling real datasets.

These differences are referred to as the domain gap, and overcoming this domain gap is one of the most important tasks in semantic segmentation, called domain adaptation. The domain adaptation task involves a source domain, where we train models on an annotated dataset, and a target domain, where the model is intended to be deployed. Ideally, the source synthetic data should substitute real data during training, with the objective of leveraging the knowledge gained from the synthetic source domain to enhance model performance in an unlabeled real target domain. 

In this work, we focus on the task of Unsupervised Domain Adaptation, where we exploit the unlabeled data of the target domain to further enhance the performance of an algorithm trained with supervision on the source domain.

Unsupervised Domain Adaptation tries to leverage data from both domains to increase the performance of a given algorithm on a target domain. This adaptation can be performed at different levels, and we can distinguish between three main trends for unsupervised domain adaptation depending on the space where the adaptation is performed \cite{10128983}: In Input Space Domain Adaptation \cite{hoyer2022daformer, hoyer2022hrda, hoyer2023mic,Tranheden2020DACSDA,9888149}, the adaptation is performed by modifying the images used to train the segmentation model. In Feature Space Domain Adaptation \cite{10260260,9796124,10.1145/3474085.3475186, wmmd, 8099590} the adaptation is performed at a feature level within the model, usually by trying to align feature distributions between domains, and in Output Space Domain Adaptation \cite{10537070,AK2023106172,8578878,vu2019advent, hoyer2022daformer} the adaptation is performed by trying to align predictions between source and target domain. 

In our work, we propose including new classes at the source domain space, without any specific knowledge of the target domain.

\subsection{Segment Anything Model}
The Segment Anything Model (SAM) \cite{kirillov2023segment} is a breakthrough in the realm of segmentation models, designed to generalize across a wide variety of segmentation tasks without requiring extensive task-specific training. SAM can use different types of input prompts, such as points, boxes, or masks, to produce zero-shot segmentation across diverse datasets, addressing the challenges posed by domain variability. By learning to segment "anything," SAM offers a robust solution that can significantly reduce the reliance on large annotated datasets, thereby accelerating the development and deployment of segmentation models across different domains. The main advantage of this model is producing high-quality semantic segmentation masks, with the drawback that no semantic information about the segmented object is provided.

\section{Method}
We propose an unsupervised and training-free pipeline (see Figure~\ref{fig:creation_pipeline}) designed to generate samples of new classes so they can be included in already existing datasets.


\subsection{Pipeline Definition.}
\begin{figure}[h]
    \centering
    \includegraphics[width=\linewidth]{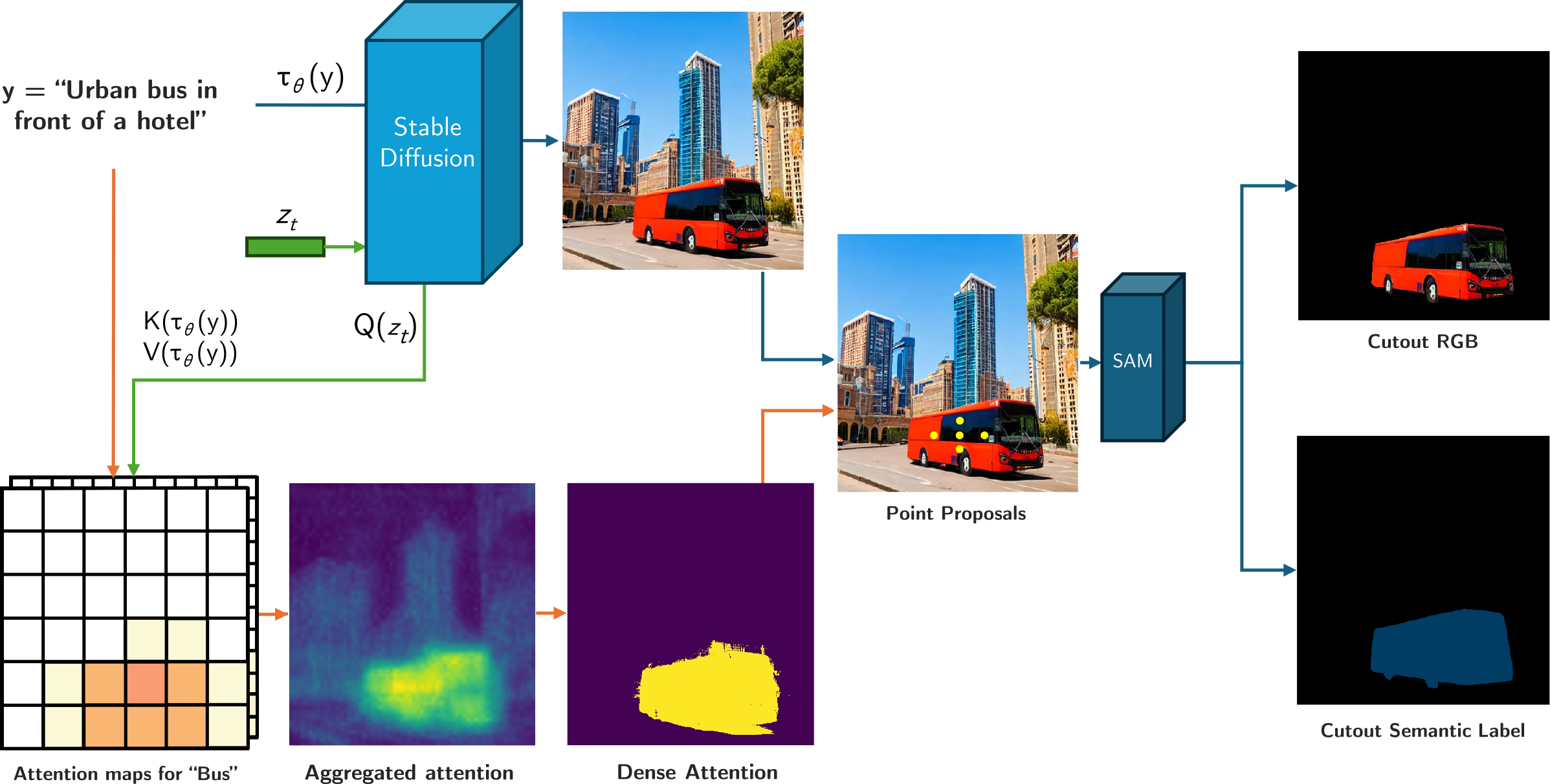}
    \caption{Schematic of our pipeline. The upper path contains the generation of the synthetic image and the lower path depicts the process of obtaining the semantic mask for the generated example.}
    \label{fig:creation_pipeline}
\end{figure}

The general scheme for our class sample generation process is shown in Figure~\ref{fig:creation_pipeline}. First, we generate a list of text prompts $\left\{\textbf{p}_j\right\}_{j=0}^N$ by randomly mixing a list of possible class types and possible locations: e.g., for the class \textit{bus}, we would have different class examples like \textit{school bus, tour bus, trolleybus}, among others, and then a list of locations for that class to ensure visual variability, i.e., \textit{in the street, at the airport, on a scenic route}, etc.

These $\textbf{p}_j$ prompts are used as text conditioning for a Stable Diffusion \cite{rombach_high-resolution_2022} model that, for a given noise vector $z_t$, generates an image $\textbf{I}_{j} = \mathcal{D}\bigl(\epsilon_\theta(z_t,t,\tau_\theta(p_j))\bigr) $, while storing the queries $Q_{i,t}$ corresponding to the $i$-spatial location at time step $t$ from the denoising process. Now, we localize the object from the new class in the image using the attention for its class token. To maximize this attention, and following the method from \cite{marcos2024open}, we use a modified version of our input prompt $\bar{p_j}$ where we simplify the class definition, trying to reduce the semantic class to a single token. For example, for our \textit{bus} example, we would change \textit{school bus} to \textit{bus}. The text embedding of this modified prompt $\tau_\theta(\bar{p_j})$ is then processed to produce the associated linear projections for its attention $\bar{K} = W_{\bar{K}}^{(i)} \cdot \tau_\theta(\bar{p_j})$, that is then combined with the pixel queries to create open-vocabulary attention matrices $A(Q_{i,t},\bar{K}_i)$. We then resize the attention matrices to a common resolution and aggregate the matrices across layers, time steps, and attention heads similarly to DAAM \cite{tang2022daam}:

\begin{equation*}
    D_{\tau_\theta(p_j),k}(\tau_\theta(\bar{p_j})) =\sum_{i,t,h}(A_{h,k}(Q_{i,t},\bar{K_i})) \in \mathbb{R} ^{W \times H}
\end{equation*}

This defines the attention matrix for each $k$-th token. We then select the matrix for our new class token, $\textbf{M}_q = D_{\tau_\theta(p_j),Q}$.

From this matrix, we filter out the pixels with attention values below a given threshold, $T_{th}$ which we set at 0.5. After this filter, we use a DenseCRF \cite{krahenbuhl2011efficient} to post-process the filtered attention map and obtain a dense binary mask, which is then used to obtain a bounding box that gives the location of the object from the class in the generated image \textbf{$I_j$}.

We use this bounding box to propose five points placed around the bounding box center and use them to prompt SAM\cite{kirillov2023segment} to obtain the $M_j$ binary mask for our object, which we then apply to the generated image \textbf{$I_j$} to produce the RGB cutout and semantic segmentation ground truth for the new class. 

Using variations of the automatically generated prompts and different latent vectors $z_t$, we can repeat this process multiple times to obtain a large set of RGB cutouts $\textbf{x}_{q}$ and their semantic masks $\textbf{y}_{q}$.

\subsection{Mask Curation}\label{sec:mask_curation}

\begin{figure}[h]
    \centering
    \includegraphics[width=\linewidth]{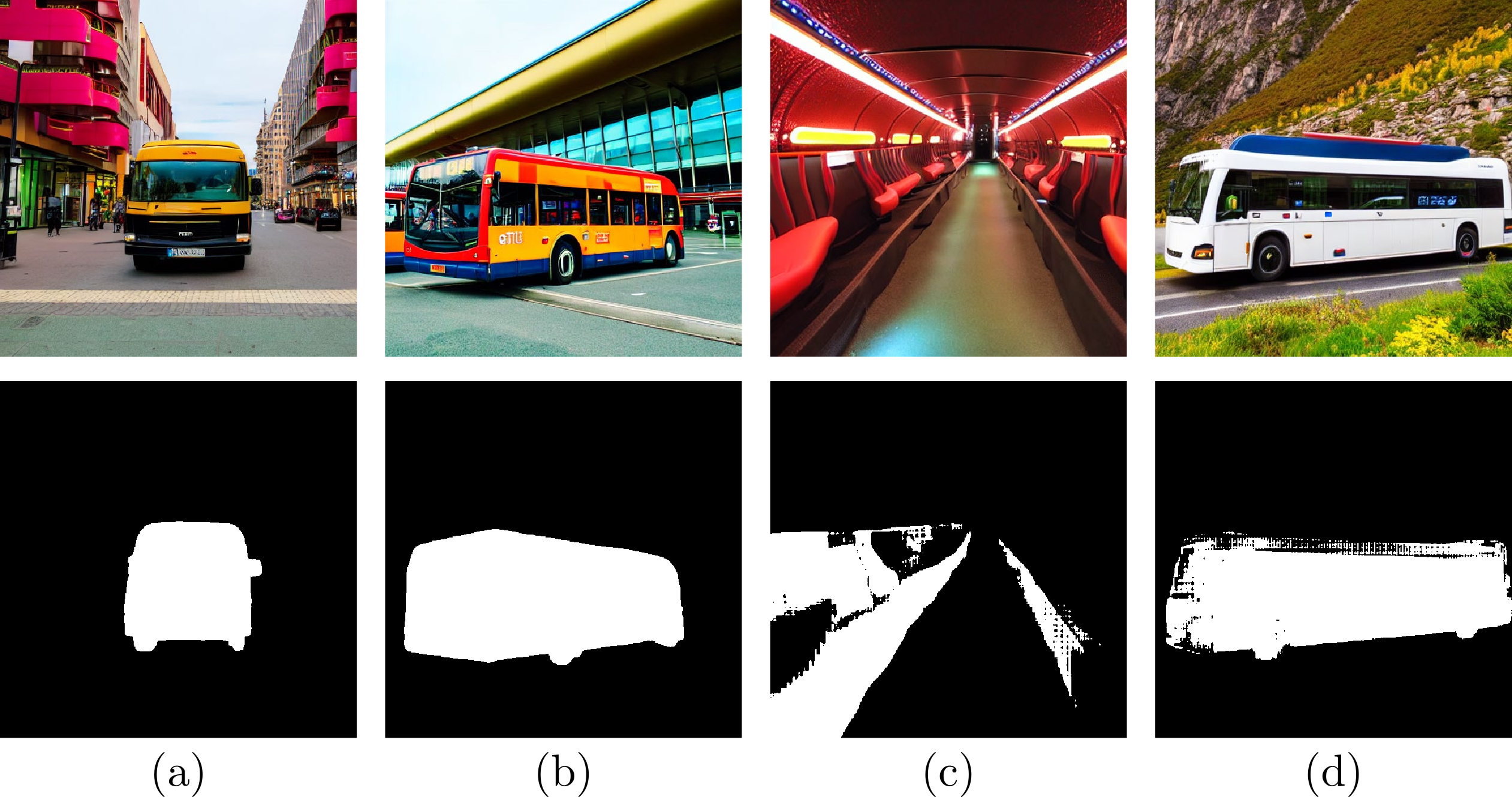}
    \caption{Images \textit{a} and \textit{b} show examples of valid images and masks for the \textit{bus} class; \textit{c} shows what seems to be the interior of a bus, and \textit{d} has a noisy mask, so both are discarded. }
    \label{fig:gen_examples}
\end{figure}

With this method, we now have multiple cutout examples for the new class we want to include in the dataset, but some of them may have some issues. Figure~\ref{fig:gen_examples} shows some examples of generated images and class masks. For example, when the attention mask occupies a notable proportion of the image, we can assume either it is a close-up, or it is highly probable that the image is not something we want (for example, image (c) in Figure~\ref{fig:gen_examples} resembles the interior of a bus). We discard these images just by setting a fixed threshold for the ratio of attention pixels which we set at 40\% of the image.

The (d) example from Figure~\ref{fig:gen_examples} shows one of the drawbacks of SAM: it may produce noisy segmentation masks, particularly on object edges, as SAM is trained on 1024x1024 images, and the resolution of the images we generate is 512x512. To remove these samples, we include an additional filtering process relying on three different metrics to filter these noisy masks: the Polsby-Popper method for measuring mask compactness, a mask contour smoothness metric, and measuring angular change along the mask of the contour metric. 

The Polsby-Popper \cite{polsby1991third,cox1927method} method is defined as:

\begin{equation}
    PP(m) = \frac{4\pi * A(m)}{P(m)^2}
\end{equation}
    
Where \textit{m} is the mask generated by SAM, $A(m)$ is the area of the mask, and $P(m)$ is the perimeter of the mask. We only keep masks where $PP(m) > 0.6$.
Masks that pass this initial filter are then processed using a perimeter smoothness metric. 
\begin{equation}
S(m) = \frac{P(m)}{P_s(m)}
\end{equation}
Where $P_s(m)$ is the perimeter obtained after smoothing the mask. If $S(m) < 1.0 $ the mask is discarded.

Finally, we process the mask using an energy metric that measures the angular change along the perimeter contour, defined as:
\begin{equation}
    E_P = \sum_{i = 1}^{N-1} \Delta \theta_i
\end{equation}

Where $\theta_i$ is the angle between two consecutive segments in the contour, so the energy for two consecutive angles would be $ \Delta \theta_i = | \theta_{i+1} - \theta_{i} |$. A high energy value indicates more irregular or noisy boundaries, with larger or more frequent changes in direction. We keep masks where $E_P < 50$. Although we could normalize this energy using the perimeter length, we found it often resulted in bigger masks with smaller noisy portions not being discarded. Thresholds were selected after manual observation of a small subset of 20 images for the \textit{bus} class and were kept the same for other classes.

\subsection{Including Novel Classes in Unsupervised Domain Adaptation Pipelines for Semantic Segmentation.}


To validate our pipeline, we will include our synthesized cutouts in an unsupervised domain adaptation (UDA) framework. In UDA we have two sets of images: A \textit{source} domain $\mathcal{X}_S = \left\{(\textbf{x}_s^i, \textbf{y}_s^i)\right\}_{i=0}^{N_s}$ composed of images with ground-truth semantic labels $\textbf{y}_r^i \in \left\{1,2,...,C\right\}^{H \times W}$, and a \textit{target} domain $\mathcal{X}_T = \left\{(\textbf{x}_t^i)\right\}_{i=0}^{N_t}$ with images but not labels, to leverage the source data to train an algorithm that effectively generalizes to the target domain. Usually, this is achieved by doing supervised training on the source data, and then using a teacher model or even the same model to generate pseudo-labels for the target data \cite{hoyer2022daformer, hoyer2022hrda, hoyer2023mic, wang2023pseudo}, which are then used in the loss computation for target images.

To include our novel class examples, we propose a method that works by modifying the data used to train on the supervised source domain. Drawing inspiration from MixUP \cite{Tranheden2020DACSDA}, we randomly combine source images $x_s^i$ and our class examples $x_q$ with a fixed probability $p_m$, by doing the dot product of the new class mask $y_q$ and the source image at a random location of the image where the class example fits entirely. With this, for a copy of the source image and labels $x_m^i = x_s^i$ and $y_m^i = y_s^i$, we randomly propose a region of the source image $roi$ where the cutout can be overlaid $x_m^i [roi]$, we include our new class sample with:

\begin{equation*}
    x_m^i[roi] = y_q^i \cdot x_q^i ,
\end{equation*}
and substitute the labels by multiplying the binary mask with the new class index:

\begin{equation*}
    y_m^i[roi] = y_q^i \cdot Q 
\end{equation*}

\begin{figure}[h]
    \centering
    \includegraphics[width=\linewidth]{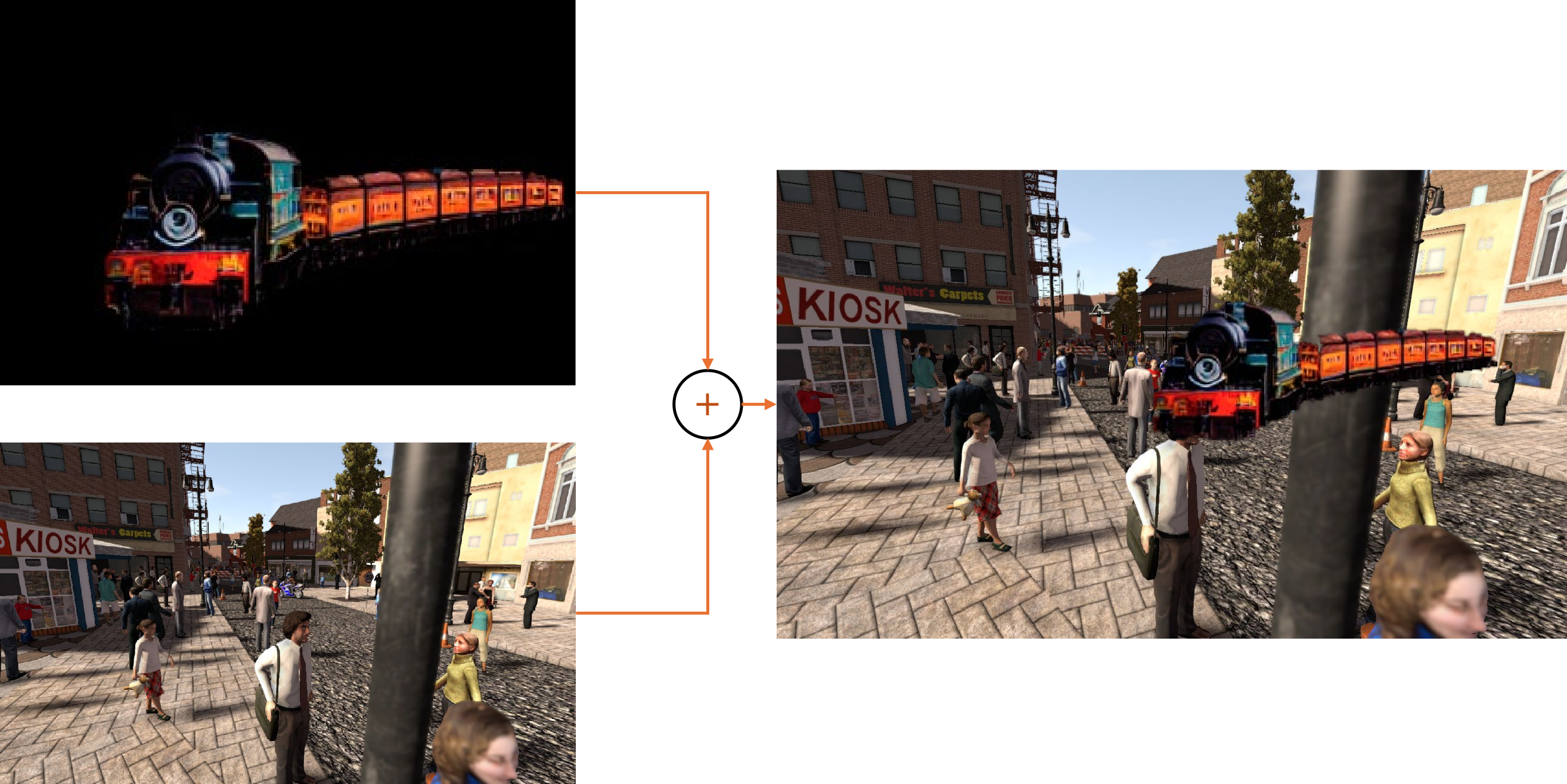}
    \caption{Example of our combination process.}
    \label{fig:diffmix}
\end{figure}

Obtaining new image-label pairs $x_m^i,y_m^i$ with $\textbf{y}_m^i \in \left\{1,2,...,C,Q\right\}^{H \times W}$, increasing the number of classes present in the labels from the source domain. Figure~\ref{fig:diffmix} shows an example of the resulting training image generated with our mix approach by combining a cutout of the class \textit{train} generated with our pipeline and an image from Synthia\cite{Ros2016synthia} dataset. For simplicity and consistency with DAFormer's \cite{hoyer2022daformer} MixUp\cite{Tranheden2020DACSDA} implementation, we do not perform augmentation on the cutouts. 

\begin{table*}[h]
    \setlength{\tabcolsep}{1pt}
    \centering
    \resizebox{\linewidth}{!}{%
    \begin{tabular}{*{20}{>{\centering\arraybackslash}p{0.8cm}}l*{1}{>{\centering\arraybackslash}p{1cm}}r*{1}{>{\centering\arraybackslash}p{0.5cm}}}
        \textbf{\quad}&\rotatebox{60}{\textbf{New Class}}& \rotatebox{60}{road} & \rotatebox{60}{sidewalk} & \rotatebox{60}{building} & \rotatebox{60}{wall} & \rotatebox{60}{fence} & \rotatebox{60}{pole} & \rotatebox{60}{traffic light} & \rotatebox{60}{traffic sign} & \rotatebox{60}{vegetation} &  \rotatebox{60}{sky} & \rotatebox{60}{person} & \rotatebox{60}{rider} & \rotatebox{60}{car} & \rotatebox{60}{\textit{truck}} & \rotatebox{60}{\textit{bus}} & \rotatebox{60}{\textit{train}} & \rotatebox{60}{motorcycle} & \rotatebox{60}{bicycle} & \rotatebox{60}{\textbf{mIoU}} \\ 
        \midrule
        \parbox[t]{2mm}{\multirow{4}{*}{\rotatebox[origin=c]{90}{Synthia}}} & - & 82.4 & 37.7 & \textbf{88.7} & 43.0 & \textbf{8.4} & \textbf{50.8} & \textbf{55.7} & 55.1 & 86.0 & 88.1 & 74.2 &\textbf{ 49.5} & 87.8 & - & 63.2 & - & 54.5 & \textbf{62.8} & 54.9 \\
         & Train & \textbf{87.5} & \textbf{50.2} & 88.4 & 44.4 & 1.6 & 49.2 & 53.1 & 50.7 & 85.3 & 92.8 & \textbf{74.5} & 48.2 & 85.9 & - & \textbf{70.6} &\textit{ 29.7} & 53.4 & 60.1 & 57.0\\
         & Truck & 82.5 & 40.8 & 88.8 & \textbf{44.6} & 6.8 & 50.4 & 55.5 & 51.0 & 85.1 & 91.7 & 67.1 & 47.6 & \textbf{90.5} & \textbf{\textit{64.0}} & 60.3 & - & \textbf{55.8} & 62.2 & 58.0 \\
         & Both & 86.5 & 47.1 & 88.3 & 44.4 & 4.4 & 49.9 & 54.1 & \textbf{54.8} & \textbf{86.6} & \textbf{93.0} & 73.1 & 41.2 & 86.3 & \textit{38.4} & 49.2 & \textbf{\textit{52.0}} & 53.5 & 60.8 & \textbf{59.1} \\ \midrule
        \parbox[t]{2mm}{\multirow{4}{*}{\rotatebox[origin=c]{90}{4AGT}}} & - & 91.5 & 68.5 & 89.1 & 43.4 & 30.1 & 50.1 & 48.4 & 59.8 & 88.3 & 92.7 & 70.7 & 35.4 & 88.0 & 25.7 & - & - & 49.9 & \textbf{61.9} & 55.2 \\
         & Train & \textbf{96.1} & 70.1 & 88.9 & 44.4 & 29.9 & 50.2 & 54.3 &\textbf{62.3} & 88.1 & 92.7 & 69.9 & 41.9 & 86.9 & \textbf{66.1} & - & \textit{42.7} & 54.8 & 58.4 & 61.0\\
         & Bus & 96.0 & 70.7 & \textbf{89.2} & \textbf{44.6} & \textbf{33.8} & \textbf{51.8} & \textbf{54.4} & 60.6 & \textbf{88.6} & \textbf{93.7} & 69.6 & 38.6 & \textbf{90.1} & 56.7 & \textit{49.5} & - & 52.0 & 60.3 & 61.1 \\
         & Both & 95.9 &\textbf{70.5} & 87.5 & 33.7 & 25.9 & 51.0 & 53.0 & 57.6 & 88.3 & 93.2 & \textbf{70.4} & \textbf{43.1} & 85.5 & 35.4 & \textbf{65.2} & \textbf{\textit{65.5}} & \textbf{\textit{55.1}} & 61.0 & \textbf{63.2}\\

        \bottomrule
    \end{tabular}}
    \caption{Per-Class performance in the DAFormer UDA pipeline. Classes introduced with our approach in \textit{italic}. (Synthetic  $\rightarrow$ Cityscapes)}
    \label{tab:new_classes}
\end{table*}

\section{Experiments}
All our tests were performed using the DAFormer\cite{hoyer2022daformer} pipeline for UDA. All models were trained for 40000 iterations, with a batch size of 2, using standard DAFormer settings.

We rely on Cityscapes \cite{Cordts2016Cityscapes} to measure our method's performance after training on two different synthetic datasets: \textcolor{red}{CARLA-4AGT} \cite{}, which  contains 16 out of 19 of Cityscapes classes, missing \textit{bus}, \textit{train} and \textit{terrain}; and Synthia Dataset\cite{Ros2016synthia}, which contains 16 out of 19 Cityscapes classes, missing  \textit{train}, \textit{truck} and \textit{terrain}.

\begin{figure}[h]
    \centering
    \includegraphics[width=.9\linewidth]{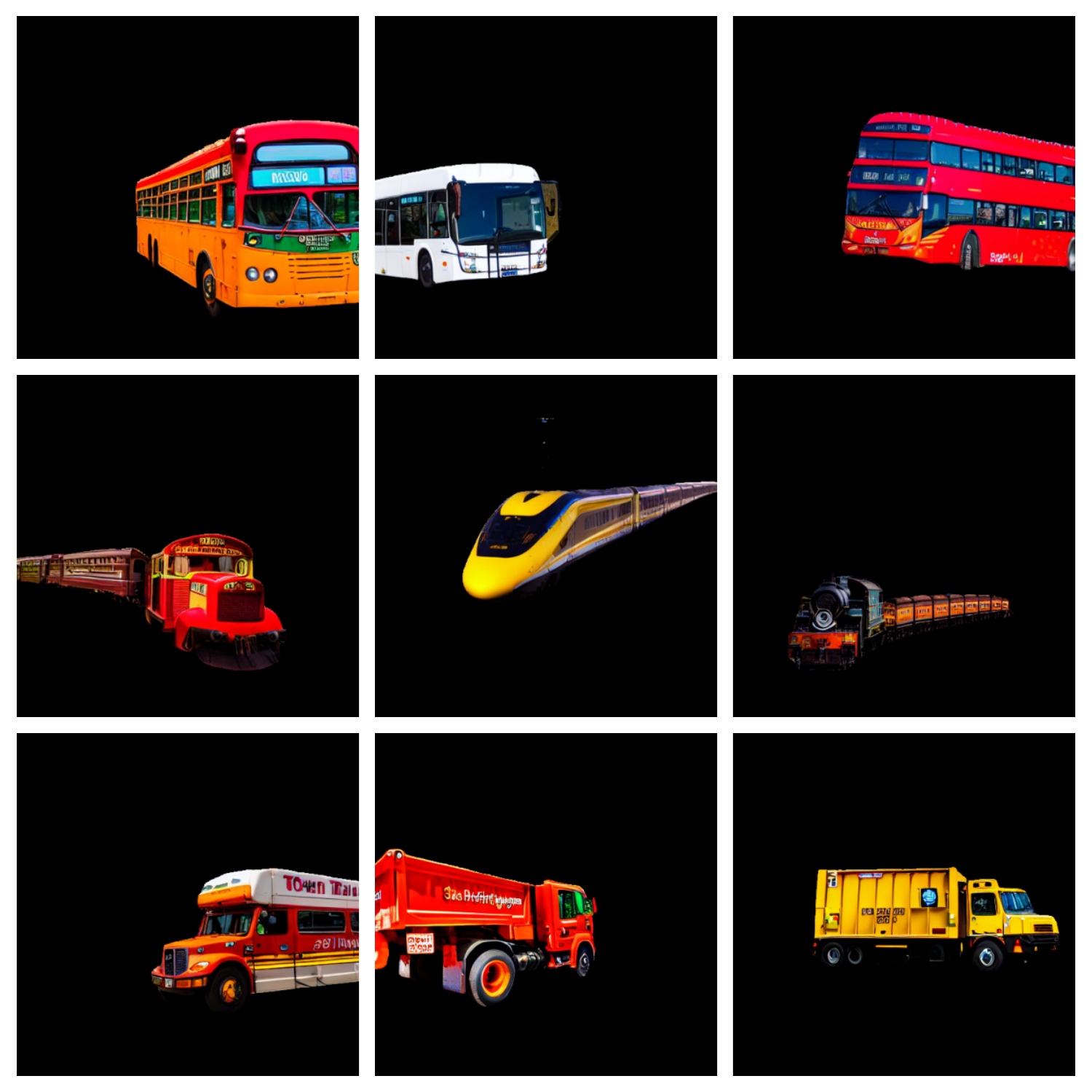}
    \caption{Each row shows cutout examples for \textit{bus},\textit{train} and \textit{truck} in descending order.}
    \label{fig:examples}
\end{figure}

For our tests, we have generated samples of \textit{bus}, \textit{truck}, and \textit{train} classes. To create the cutouts for these classes we started by prompting Llama 2 \cite{touvron2023llama} for a list of possible types of vehicles for each class, and another list of possible locations. Additionally, in the prompt, we included the styling words \textit{ego camera} and \textit{color}, and included a negative prompt with the content \textit{grayscale, artistic, painting} to maximize the output of images that resemble photographies. With these prompts, we generate images for all three classes, using our pipeline to extract cutouts and masks, and our filtering process. Figure~\ref{fig:examples} shows some filtered cutout examples for the different \textit{bus}, \textit{train}, and \textit{truck} classes.

In all our tests, we measure performance by calculating the per-class intersection over union (IoU) \cite{Everingham10thepascal}, between model predictions and the ground-truth labels pf Cityscapes validation set, with IoU defined as $\mathrm{IoU} = \frac{TP}{TP+FP+FN}$, where $TP$ and $ FP$ mean \textit{true positives} and \textit{false positives} pixels respectively, and $FN$ being the amount of false negative pixels. 

\subsection{Including New Classes on Datasets}

For this experiment, we generate 2000 cutouts for each class in \textit{bus}, \textit{truck}, and \textit{train}, and include them into source images using our proposed mix method.

In Table~\ref{tab:new_classes}, we show the results after the inclusion of the classes each dataset was missing compared to Cityscapes, both individually and at the same time. Including our cutouts is not only useful for learning a class that was previously missing from the training data, allowing to effectively segment these novel classes, but also results in performance benefits for other classes.

\begin{table*}[!b]
    \setlength{\tabcolsep}{1pt}
    \centering
    \resizebox{\linewidth}{!}{%
    \begin{tabular}{*{3}{>{\centering\arraybackslash}p{1cm}}l*{19}{>{\centering\arraybackslash}p{0.8cm}}l*{1}{>{\centering\arraybackslash}p{1cm}}r*{1}{>{\centering\arraybackslash}p{0.5cm}}}
        \textbf{\quad}&Class& \rotatebox{60}{Filtering}&\rotatebox{60}{road} & \rotatebox{60}{sidewalk} & \rotatebox{60}{building} & \rotatebox{60}{wall} & \rotatebox{60}{fence} & \rotatebox{60}{pole} & \rotatebox{60}{traffic light} & \rotatebox{60}{traffic sign} & \rotatebox{60}{vegetation} &  \rotatebox{60}{sky} & \rotatebox{60}{person} & \rotatebox{60}{rider} & \rotatebox{60}{car} & \rotatebox{60}{\textit{truck}} & \rotatebox{60}{\textit{bus}} & \rotatebox{60}{\textit{train}} & \rotatebox{60}{motorcycle} & \rotatebox{60}{bicycle} & \rotatebox{60}{\textbf{mIoU}} \\ 
        \midrule
        \parbox[t]{2mm}{\multirow{4}{*}{\rotatebox[origin=c]{90}{Synthia}}} & \multirow{2}{*}{Train} & \xmark & 85,9 & 45,6 & 88,5 & 45,7 & 8,3 & 50,1 & 54,3 & 48,8 & 86,8 & 90,6 & 73,1 & 40,2 & 89,4 & - & 57,3 & \textit{0.0} & 49 & 54,3 & 53,8 \\
         & & \cmark & 87.5 & 50.2 & 88.4 & 44.4 & 1.6 & 49.2 & 53.1 & 50.7 & 85.3 & 92.8 & 74.5 & 48.2 & 85.9 & - & 70.6 & \textit{\textbf{29.7}} & 53.4 & 60.1 & 57.0\\ \cmidrule{2-22}
          & \multirow{2}{*}{Truck} & \xmark & 85.4 & 43.5 & 89.1 & 47.4 & 9.2 & 49.9 & 54.5 & 56.1 & 86.4 & 87.9 & 69.0 & 43.1 & 88.9 & \textit{61.2} & 51.2 & - & 53.2 & 61.3 & 57.6 \\
         &  & \cmark & 82.5 & 40.8 & 88.8 & 44.6 & 6.8 & 50.4 & 55.5 & 51.0 & 85.1 & 91.7 & 67.1 & 47.6 & 90.5 & \textbf{\textit{64.0}} & 60.3 & - & 55.8 & 62.2 & 58.0 \\ \midrule
        \parbox[t]{2mm}{\multirow{4}{*}{\rotatebox[origin=c]{90}{4AGT}}} & \multirow{2}{*}{Train} & \xmark & 96,1 & 64,5 & 88,8 & 41,2 & 27,5 & 50,2 & 53,1 & 59,8 & 88,2 & 92,8 & 69,9 & 41,9 & 88,3 & 26,4 & - & \textit{13,2} & 51,2 & 62,1 & 56,4 \\
         & &\cmark & 96.1 & 70.1 & 88.9 & 44.4 & 29.9 & 50.2 & 54.3 &62.3 & 88.1 & 92.7 & 69.9 & 41.9 & 86.9 & 66.1 & - & \textit{\textbf{42.7}} & 54.8 & 58.4 & 61.0\\  \cmidrule{2-22}
         & \multirow{2}{*}{Bus} & \xmark & 95,4 & 70,3 & 88,7 & 44,2 & 33,2 & 52,3 & 52,1 & 61,3 & 88 &  93,4 & 68,3 & 44,4 & 87,5 & 47,0 & \textit{28,3} & - & 49,1 & 63,2 & 59,3 \\ 
         & &\cmark & 96.0 & 70.7 & 89.2 & 44.6 & 33.8 & 51.8 & 54.4 & 60.6 & 88.6 & 93.7 & 69.6 & 38.6 & 90.1 & 56.7 & \textbf{\textit{49.5}} & - & 52.0 & 60.3 & 61.1 \\ 
        
        \bottomrule
    \end{tabular}}
    \caption{Performance comparison when training with and without mask filtering. Novel classes in italic. (Synthetic  $\rightarrow$ Cityscapes)}
    \label{tab:comparison_other_methods}
\end{table*}

 \begin{figure}
     \centering
     \includegraphics[width=0.95\linewidth]{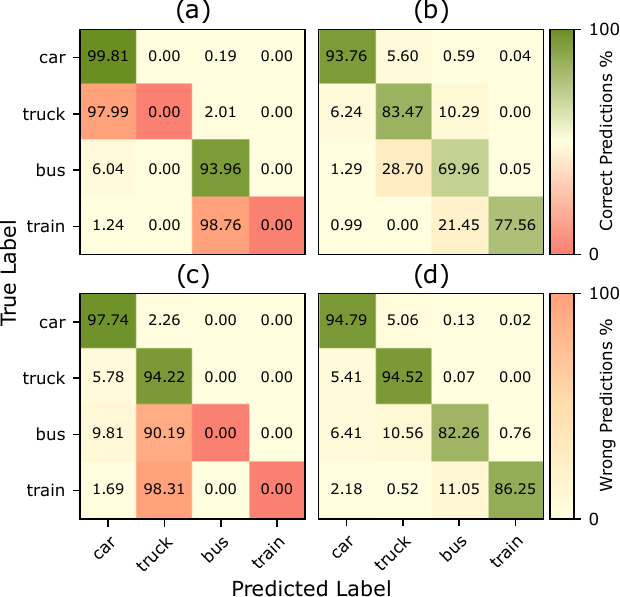}
     \caption{Confusion matrices for vehicle classes before and after including the two missing classes. (a) Synthia baseline (b) Synthia + our approach, (c) CARLA-4AGT (d) CARLA-4AGT + our approach}
     \label{fig:confusion_matrices}
 \end{figure}
 
This performance uplift for other classes is more evident when looking at the confusion matrices from Figure~\ref{fig:confusion_matrices}, where we can see how \textit{train} elements were being predicted as \textit{bus} and \textit{trucks} were predicted as \textit{cars} when adapting from Synthia to Cityscapes. Still, with the inclusion of our synthesized cutouts, the model is now able to segment the \textit{truck} class properly. For the model trained with the CARLA-4AGT dataset, we see a similar behavior but in this case, the \textit{bus} and \textit{train} classes are being wrongfully labeled as \textit{truck}.

\subsection{Ablation Tests} 
\begin{figure}[h]
    \centering
    \includegraphics[width=\linewidth]{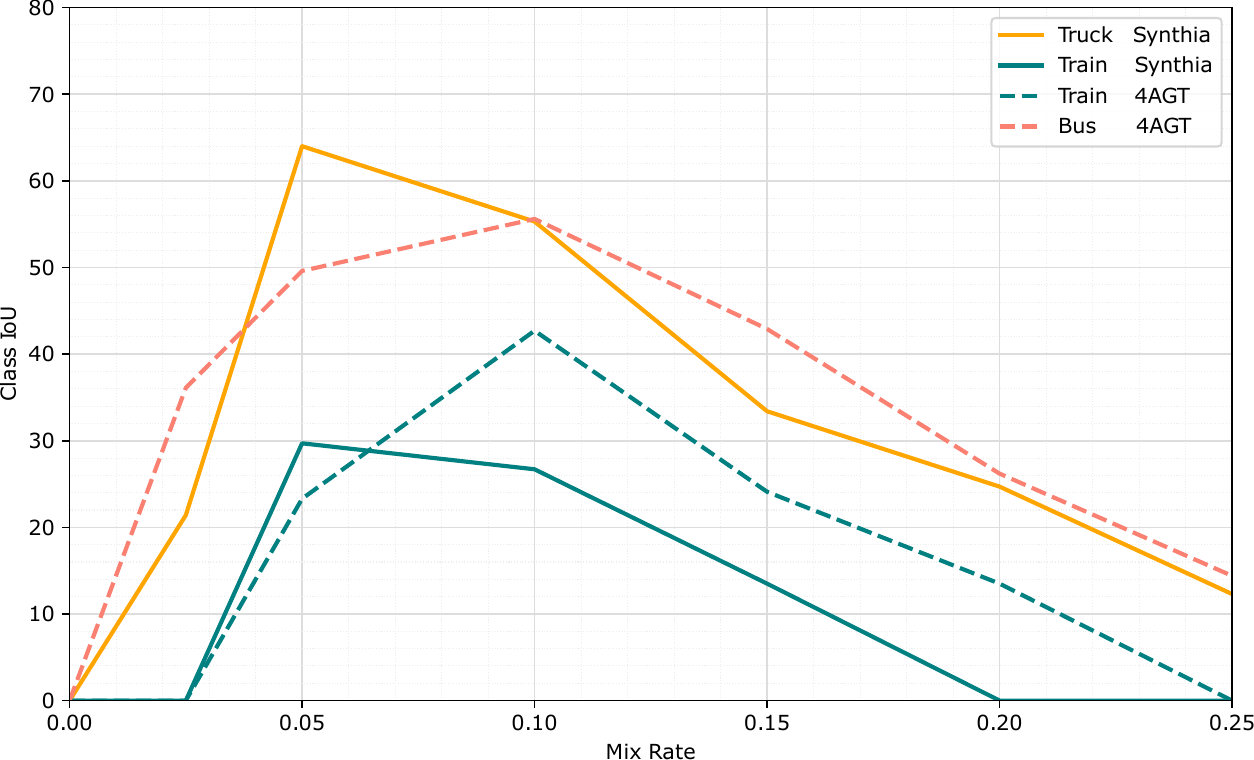}
    \caption{Performance of the \textit{bus}, \textit{truck}, and \textit{train} classes when using our approach to complete Synthia and CARLA-4AGT datasets. The optimal value varies across datasets.}
    \label{fig:probability_ablation}
\end{figure}

\paragraph{Impact of Appearance Rate.}We perform an ablation test for different values of the $p_m$ parameter, that decides the probability of a novel class to be inserted into an image. In Figure~\ref{fig:probability_ablation} we show the per-class performance of new classes depending on the mixing probability parameter. If the parameter is too high, the model overfits and loses generalization capabilities. If the parameter is too low, the number of examples seen is not high enough, and the model is not able to learn to segment the new class. We find that the optimal parameter value varies across datasets. We believe this is due to the differences in image sizes between Synthia (1280x760) and CARLA-4AGT (2048x1024), which results in the cutouts occupying a smaller region of the image for CARLA-4AGT images, so when performing the 512x512 crop during training, there is a smaller probability of sampling the cutout, so a higher $p_m$ value compensates for the smaller chance of being cropped.

\vspace{-5mm}
\paragraph{Mask Filtering Evaluation.}
To evaluate the impact of the cutout filtering described in Section \ref{sec:mask_curation} we now train the same DAFormer setup using 2000 unfiltered cutouts for \textit{bus}, \textit{truck}, and \textit{train} classes, and compare them to the results after training on 2000 filtered images. Results are summarized in Table~\ref{tab:comparison_other_methods}. For all classes, there was a performance degradation when not using mask filtering, and we can see how some classes seem to be slightly affected by not filtering cutouts (Synthia+Truck), while other classes suffer a noticeable drop in performance, also affecting the overall model performance, for example, when training Synthia including unfiltered train cutouts, we observe a global mIoU performance drop compared to the baselines (53.8 vs 54.9). Novel-class performance degradation when using unfiltered cutouts is aligned with the amount of generated samples required to obtain the 2000 valid cutouts after filtering: For the \textit{truck} class, we had to generate ~2600 images and masks to reach 2000 \textit{valid} cutouts, while the \textit{train} class, that showed more performance degradation, required ~3700 generations. This suggests filtering process is necessary to discard samples that hinder the performance of models upon training with them.

\section{Conclusions}
In this work, an automatic pipeline that leverages Stable Diffusion to generate synthetic class examples that can be used to train semantic segmentation algorithms, and propose a method for exploiting this pipeline by including novel classes in existing semantic segmentation datasets to extend their semantic categories. We show results in unsupervised domain adaptation pipelines proving that our method successfully enables learning classes not originally available in the source datasets with a performance similar to that obtained for other existing classes.

{\small
\bibliographystyle{ieee_fullname}
\bibliography{egbib}
}

\end{document}